\documentclass[11pt]{article}

\usepackage[final]{acl}

\usepackage{times}
\usepackage{latexsym}
\usepackage{flushend}

\usepackage[T1]{fontenc}

\usepackage[utf8]{inputenc}

\usepackage{microtype}

\usepackage{inconsolata}

\usepackage{graphicx}

\usepackage{booktabs}
\usepackage{array}
\usepackage{colortbl}
\usepackage{xcolor}
\usepackage{makecell}
\usepackage{tabularx}
\usepackage{tikz}
\usepackage[T1]{fontenc}

\definecolor{dotGreen}{HTML}{1D9E75}
\definecolor{dotAmber}{HTML}{BA7517}
\definecolor{dotGray}{HTML}{D3D1C7}
\definecolor{dotGrayText}{HTML}{5F5E5A}
\definecolor{badgeGrayBg}{HTML}{D3D1C7}
\definecolor{badgeBlueBg}{HTML}{B5D4F4}
\definecolor{badgeGrayFg}{HTML}{444441}
\definecolor{badgeBlueFg}{HTML}{0C447C}
\definecolor{sectionFg}{HTML}{888884}

\newcommand{\mydot}[3]{%
  \tikz[baseline=-0.55ex]\node[circle,fill=#1,text=#2,inner sep=0pt,
    minimum size=13pt,font=\bfseries\fontsize{6.5}{6.5}\selectfont]{#3};}
\newcommand{\dotY}{\mydot{dotGreen}{white}{Y}}
\newcommand{\dotN}{\mydot{dotGray}{dotGrayText}{N}}
\newcommand{\dotP}{\mydot{dotAmber}{white}{\raise.1ex\hbox{$\sim$}}}

\newcommand{\badgegray}[1]{{\fontsize{6.5}{6.5}\selectfont
  \colorbox{badgeGrayBg}{\color{badgeGrayFg}\,#1\,}}}
\newcommand{\badgeblue}[1]{{\fontsize{6.5}{6.5}\selectfont
  \colorbox{badgeBlueBg}{\color{badgeBlueFg}\,#1\,}}}

\newcommand{\namecell}[3]{%
  \begin{tabular}[c]{@{}l@{}}
    {\small\bfseries #1}\\[-1pt]
    {\fontsize{7.5}{8}\selectfont\color{dotGrayText}#2}\\[-2pt]
    #3
  \end{tabular}%
}

\title{Position: Evaluations of AI Moral Reasoning Still Miss Half of the Picture}

\author{
  \textbf{Aidan Kierans\textsuperscript{1}},
  \textbf{Ritam Dutt\textsuperscript{2}},
  \textbf{Kaley Rittichier\textsuperscript{1}},
  \textbf{Shiri Dori-Hacohen\textsuperscript{1}},
  \textbf{Avijit Ghosh\textsuperscript{3,1}}
\\
  \textsuperscript{1}University of Connecticut,
  \textsuperscript{2}Carnegie Mellon University,
  \textsuperscript{3}Hugging Face
\\
  \small{
    \textbf{Correspondence:} \href{mailto:aidan.kierans@uconn.edu}{aidan.kierans@uconn.edu}
  }
}

\begin{document}
\maketitle
\begin{abstract}

Recent work on evaluating the moral competence of large language models (LLMs) has focused primarily on what we call the moral value problem, i.e., whether model outputs align with human moral values. In contrast, the moral norm problem, i.e., whether models can identify and correctly apply context-sensitive moral norms, remains underexplored. We posit that this imbalance stems from the field's reliance on descriptive ethics frameworks, such as Moral Foundations Theory and Kohlberg's stages of moral development, which emphasize value representation over normative application. We review existing benchmarks and evaluation methods, and show that they cluster heavily around the value problem, while discussion regarding normative ethics remains underrepresented. We identify three crucial gaps: (i) the absence of high-quality ground-truth data for moral norms and their applications, (ii) insufficient evaluation of intermediate reasoning processes, and (iii) limited attention to the identification of morally relevant features in context. Subsequently, we propose a research agenda that includes the development of standardized formal representations for normative theories, the construction of expert-annotated datasets capturing norm application, and evaluation protocols that explicitly distinguish between values-level and norms-level competence. Our goal is to encourage a more systematic study of normative reasoning in LLMs.

\end{abstract}

\section{Introduction}

Users increasingly rely on large language models (LLMs) for moral advice. Recent evidence suggests that such systems are perceived as comparable to expert ethicists in apparent moral expertise \citep{dillion_ai_2025}. Regardless of whether this trust is warranted, its prevalence makes it important to characterize what current evaluations of AI moral reasoning measure, and what they omit.

We frame the evaluation of moral reasoning in LLMs as consisting of two related but distinct problems. The first is the ``\textit{moral value problem}'', which asks whether model outputs reflect human moral values, understood as broad preferences and priorities. The second is the ``\textit{moral norm problem}'', which asks whether models can identify and correctly apply moral principles that determine how those values translate into judgments in specific contexts. While the value problem concerns alignment with observed human attitudes, the norm problem concerns the application of structured principles drawn from normative ethics.

Prior work has focused largely on the moral value problem. Empirical studies have compared LLM outputs with human responses using instruments such as the Moral Machine experiment, moral foundations questionnaires, and large-scale value surveys. These approaches align with descriptive ethics, which studies patterns in human moral beliefs and preferences. As a result, existing benchmarks primarily assess whether models reproduce distributions of human values.

In contrast, the moral norm problem has received limited attention. Addressing this problem requires engagement with normative ethics, which studies which principles are correct and how they apply in particular cases. Values alone are insufficient to determine moral judgments; norms specify how values constrain decisions in context. A model may approximate human value distributions while failing to construct valid arguments within established ethical frameworks, recognize when specific principles apply, or identify morally relevant features of novel scenarios. One reason for this gap is that normative ethics has not been systematically represented in forms amenable to computational evaluation. However, the underlying theories and principles are well-documented; the primary challenge lies in organizing them into structured representations that support benchmarking.

In this paper, we review existing approaches to evaluating moral competence in LLMs and map them onto realistic components of moral reasoning. We show that current evaluations concentrate on descriptive ethics and values-level alignment, with limited coverage of norms-level reasoning. Based on this analysis, we outline directions for developing datasets, representations, and evaluation protocols that enable systematic assessment of normative moral reasoning in AI systems.

\section{Background}

The empirical study of human moral values has produced well-established frameworks. Moral Foundations Theory (MFT) identifies a set of foundational moral concerns \footnote{ The main dimensions for MFT are care/harm, fairness/cheating, loyalty/betrayal, authority/subversion, sanctity/degradation, and the later-added liberty/oppression} that structure moral intuitions across cultures \citep{graham_moral_2013}.  Schwartz's Theory of Basic Human Values provides a complementary framework organized around dimensions such as self-transcendence, conservation, openness to change, and self-enhancement \citep{schwartz_overview_2012}. Both offer validated instruments for measuring what people value, and both have been widely adopted in the AI alignment literature as a basis for assessing model behavior.

Research on moral decision-making has also produced large-scale datasets of human judgments. The Moral Machine experiment collected responses to autonomous vehicle dilemmas at scale and identified consistent patterns in how participants trade off outcomes \citep{awad_moral_2018}. In parallel, research in moral psychology finds that human judgments draw on both outcome-based reasoning and rule-based responses, often associated with consequentialist and deontological patterns \citep{cushman_action_2013}. These results provide a basis for comparing model outputs with human decisions across controlled scenarios.

Recent work addresses alignment in settings where moral views differ across individuals or groups. Approaches to pluralistic alignment draw on social choice theory to formalize how conflicting preferences can be aggregated or represented \citep{sorensen2024roadmappluralisticalignment}. New benchmarks evaluate whether LLMs capture the distribution of moral opinions observed in human populations, rather than converging to a single response \citep{russo_pluralistic_2025, poole-dayan_benchmarking_2026}. Other studies examine consistency across related moral judgments \citep{moore_are_2024} and the effects of temporal variation in human feedback on alignment outcomes \citep{keswani_moral_2025}.

Within computational ethics \citep{tolmeijer_implementations_2021}, this line of work focuses on representing and evaluating descriptive ethics. In contrast, comparatively little work has addressed how to represent and evaluate normative ethics in computational settings. We now provide a detailed description of the state of the machine ethics evaluation below.

\section{The State of Machine Ethics Evaluations}
\label{sec:lit_review}
Recent work has begun to examine how LLMs are evaluated for moral competence. \citet{snoswell_beyond_2026} found that while a subset of papers assesses model-generated justifications, these evaluations typically rely on surface-level checks (e.g., consistency, hallucination) or subjective ratings, and do not test whether reasoning supports final decisions. They argue for decomposing moral reasoning into intermediate steps, evaluating performance against expert standards, and incorporating a broader range of normative theories beyond the descriptive frameworks that dominate current practice. Building on this observation, we organize the literature as follows: Section \ref{sec:value-problem} surveys benchmarks that target the \emph{moral value problem}, Section \ref{sec:norm-problem} surveys the smaller set targeting the \emph{moral norm problem}, and Section \ref{sec:conflation} diagnoses a recurring failure mode, i.e. the use of value-level instruments as proxies for norm-level competence, that emerges when these two strands are not clearly separated. Figure \ref{fig:placeholder} summarizes the benchmarks discussed across both subsections along different dimensions.

\begin{figure*}[!t]
    \noindent{\large\bfseries What current AI morality evaluations miss}\\[1pt]
    
    \vspace{0.5em}
    
    \resizebox{\textwidth}{!}{%
    \noindent\begin{tabularx}{\textwidth}{%
      >{\raggedright\arraybackslash}p{5.4cm}
      *{5}{>{\centering\arraybackslash}X}
    }
    \toprule
    & \multicolumn{2}{c}{\fontsize{7}{7}\selectfont\color{sectionFg}\MakeUppercase{Construct coverage}}
    & \multicolumn{3}{c}{\fontsize{7}{7}\selectfont\color{sectionFg}\MakeUppercase{Evaluation infrastructure (\S \ref{sec:gaps_in_cirrent_approaches})}} \\[1pt]
    \cmidrule(lr){2-3}\cmidrule(lr){4-6}
    &
    \makecell{\fontsize{7.5}{9}\selectfont Moral values\\\fontsize{7.5}{9}\selectfont assessed} &
    \makecell{\fontsize{7.5}{9}\selectfont Normative\\\fontsize{7.5}{9}\selectfont theories engaged} &
    \makecell{\fontsize{7.5}{9}\selectfont Shared norm\\\fontsize{7.5}{9}\selectfont vocabulary} &
    \makecell{\fontsize{7.5}{9}\selectfont Reasoning trace\\\fontsize{7.5}{9}\selectfont evaluated} &
    \makecell{\fontsize{7.5}{9}\selectfont Moral salience\\\fontsize{7.5}{9}\selectfont identification} \\
    \midrule
    
    \multicolumn{6}{l}{\fontsize{7.5}{9}\selectfont\color{sectionFg}\MakeUppercase{Descriptive ethics / values frameworks}}\\[-1pt]
    \midrule
    
    \namecell{Moral Machine}{\cite{awad_moral_2018}}{\badgegray{Dilemma judgments}}
    & \dotY & \dotN & \dotN & \dotN & \dotN \\[2pt]
    
    \namecell{ETHICS Dataset}{\cite{hendrycks2021ethics}}{\badgegray{Moral concept judgements}}
    & \dotP & \dotY & \dotN & \dotN & \dotP \\[2pt]
    
    \namecell{MFT Moral Hypocrisy}{\cite{nunes_are_2024}}{\badgegray{MFQ + MFV questionnaires}}
    & \dotY & \dotN & \dotN & \dotN & \dotN \\[2pt]
    
    \namecell{LLM Ethics Benchmark}{\cite{jiao_llm_2025}}{\badgegray{MFT + Kohlberg stages}}
    & \dotY & \dotN & \dotN & \dotN & \dotN \\[2pt]
    
    \midrule
    
    \multicolumn{6}{l}{\fontsize{7.5}{9}\selectfont\color{sectionFg}\MakeUppercase{Normative ethics / reasoning-focused}}\\[-1pt]
    \midrule
    
    \namecell{Policy-based deliberation}{\cite{rao_ethical_2023}}{\badgeblue{Theory-specific policies}}
    & \dotN & \dotP & \dotP & \dotP & \dotP \\[2pt]
    
    \namecell{Theory-lens reasoning}{\cite{zhou_rethinking_2024}}{\badgeblue{Direct theory application}}
    & \dotP & \dotP & \dotN & \dotP & \dotN \\[2pt]
    
    \namecell{MoralLens}{\cite{samway_are_2025}}{\badgeblue{16-rationale taxonomy}}
    & \dotP & \dotY & \dotN & \dotY & \dotP \\[2pt]
    
    \namecell{MoReBench}{\cite{chiu_morebench_2025}}{\badgeblue{Criterion-fulfillment rubric}}
    & \dotP & \dotY & \dotN & \dotY & \dotP \\[2pt]
    
    \namecell{PhilosophyBench}{\cite{philosophy_bench_2026}}{\badgeblue{Clustered rationale classification}}
    & \dotN & \dotP & \dotN & \dotY & \dotN \\[2pt]

    \bottomrule
    \end{tabularx}
    }
    
    \vspace{0.4em}
    
    \noindent
    \dotY\;{\footnotesize Addressed}
    \hspace{1.2em}
    \dotP\;{\footnotesize Partial / ad hoc}
    \hspace{1.2em}
    \dotN\;{\footnotesize Not addressed}
    
    \vspace{0.25em}

    \caption{Coverage across representative benchmarks, organized by the dimensions identified in \S \ref{sec:lit_review}--\ref{sec:gaps_in_cirrent_approaches}. The audit surfaces three gaps that current benchmarks share. We divide the benchmarks into two broad categories based on whether they score based on alignment to human judgment data or theory-based reasoning quality. We were not able to find any existing benchmark that fulfills all of our Evaluation Infrastructure criteria.
    }
    \label{fig:placeholder}
\end{figure*}

\subsection{The Moral Value Problem: What Does the AI Care About?}
\label{sec:value-problem}

A common approach to evaluating moral competence in LLMs is to test whether model outputs reflect human moral values. This is typically operationalized through multiple-choice value questionnaires, dilemma-based tasks (e.g., trolley problems), or domain-specific scenarios such as those in medical ethics \citep{soffer_disagreements_2024}. For example, \citet{nunes_are_2024} administer both the Moral Foundations Questionnaire and the Moral Foundations Vignettes to LLMs. They find that models exhibit internal consistency within each instrument but produce conflicting responses when abstract value endorsements are compared with judgments about concrete violations. Other work shows that generative settings can still reveal model priorities in cases where values conflict, by analyzing responses to value trade-off scenarios \citep{liu_generative_2026}.

This line of work aligns with a broader effort to characterize model behavior using tools from psychology. Evaluating values-level alignment is relatively straightforward: researchers adapt an existing instrument, apply it to both human participants and models, and compare the resulting distributions. Open questions remain about dataset selection, aggregation across populations, and cross-cultural coverage, but these are methodological challenges within an established paradigm. In the terminology of computational ethics \citep{tolmeijer_implementations_2021}, this corresponds to formalizing descriptive ethics and evaluating machine behavior against it.

\subsection{The Moral Norm Problem: Can the AI Apply Moral Principles?}
\label{sec:norm-problem}

The moral norm problem concerns whether LLMs can identify and apply the principles that determine how values should guide decisions in specific contexts. Within computational ethics, this requires formalizing normative ethics and designing evaluations that test principle application rather than value representation.

Only a limited number of benchmarks address this problem directly. MoralLens \citep{samway_are_2025} and PhilosophyBench \citep{philosophy_bench_2026}, for example, evaluate whether model reasoning aligns with rationales linked to consequentialist and deontological theory. \citet{rao_ethical_2023} introduce a policy-based framework in which sets of theory-linked rules are used to guide and assess in-context ethical reasoning. Related work examines whether models can apply established moral theories to novel scenarios \citep{zhou_rethinking_2024}.

Across these approaches, a common limitation is the lack of shared datasets that map normative theories to general principles or fine-grained rules. As a result, each benchmark constructs its own set of theory-derived norms. This limits comparability across studies and prevents cumulative progress: new benchmarks do not build on prior resources, and results cannot be evaluated against a common standard.

\subsection{The Values-Norms Conflation}
\label{sec:conflation}

More formally, the value
problem is the task of predicting, for a moral question $q$, the
distribution $P(j \mid q, \pi)$ of judgments $j$ that a population
$\pi$ would produce.  The norm problem is the task of producing, for a
question $q$ and a normative theory $T$, a judgment $j$ together with
a justification $r$ such that $r$ is a valid derivation of $j$ from
the principles of $T$ applied to the morally relevant features of
$q$. The two are independent: a model may match $P(j \mid q, \pi)$
without producing any valid $r$, and a model may produce valid $r$
under $T$ while diverging from $P(j \mid q, \pi)$ when $\pi$
disagrees with $T$.

A recurring pattern in past literature is that instruments that measure moral values are conflated with those that measure moral norms.  A prime example is the adoption of  Moral Foundations Theory (MFT) to make claims about norm-level competence. MFT~\citep{graham_moral_2013} provides a structured account of which moral concerns are salient to a
respondent, but it is descriptive by design: it characterizes what
people tend to care about, not how those concerns should be weighed
or what they license in a given situation. Treating an MFT-based
measurement as evidence of normative reasoning therefore substitutes
one construct (value endorsement) for another (norm application).

To illustrate, consider the dimension of \textit{care/harm} as outlined in MFT. MFT measures the degree to which a respondent treats harm-related cues as morally salient, but ``caring about harm'' is consistent with radically different normative prescriptions. A utilitarian aggregates harms across all affected parties; a Kantian treats the prohibition on using persons as mere means as inviolable regardless of aggregate harm; a virtue ethicist asks what a compassionate agent would do. A model that scores high on the care foundation has shown only that it treats harm as morally salient, divorced from whether the model can apply any of these principles to determine which action that salience licenses. The same holds for the \textit{fairness/cheating} dimension, where equality, equity, proportionality, and procedural justice all activate the fairness foundation but prescribe different verdicts in distributive cases. Foundation-level activation thus fails to accurately determine the normative output, which is precisely what a norm-level evaluation must measure.

This pattern is partly an artifact of available tooling. MFT and related frameworks provide validated instruments and well-defined categories that port readily into computational pipelines. Normative ethics, by contrast, comprises competing theories, such as consequentialism, deontology, virtue ethics, care ethics, and contractualism, without standardized representations or measurement instruments. The path of least resistance is to reuse value-level instruments and relabel the output, with relabeling being the conflation.

The ``LLM Ethics Benchmark'' of \citet{jiao_llm_2025} illustrates how the conflation operates in practice. The benchmark explicitly defines moral reasoning in terms that include identifying dilemmas, weighing considerations, and applying principles to reach justified conclusions, i.e., a norm-problem specification. Its implementation, however, relies on MFT to represent both the values \emph{and} the principles, with scoring keyed to foundation-level patterns. Because MFT does not specify application criteria, the resulting measurement collapses back onto the value problem regardless of how the construct is framed in the paper's prose. Applied to Figure \ref{fig:placeholder}, the benchmark registers on \emph{moral values assessed} but not on \emph{normative theories engaged}; this pattern recurs across the other descriptive ethics benchmarks.

We frame the conflation as a \textbf{construct validity problem} in the NLP measurement literature: a benchmark's operationalization fails to track the construct it purports to measure. A model that matches human value distributions on an MFT instrument has not thereby demonstrated the ability to
identify morally relevant features, select an applicable principle,
or derive a verdict from that principle, each of which a norm-level
claim requires. Closing this gap requires evaluation infrastructure
that the field does not yet have, as we discuss below.

\section{Gaps in Current Approaches}
\label{sec:gaps_in_cirrent_approaches}
\subsection{Missing Ground Truth for Moral Norms}

A central limitation is the lack of broadly applicable ground-truth data for normative ethics, i.e., representations of the norms endorsed by different moral theories. We thus advocate for operational representations of the principles and reasoning patterns that link values to judgments under specific theories. In the absence of such representations, each benchmark constructs its own dataset of moral norms, which limits comparability across studies.

In contrast, descriptive ethics benefits from well-established datasets and measurement tools. There is no analogous infrastructure for normative ethics. For a given normative theory, existing resources fail to systematically encode the principles it endorses, how those principles apply across contexts, or what constitutes correct application.

Some prior work provides partial foundations. \citet{hammerton_fundamental_2025} formalize moral theories in terms of the abstract properties they prioritize, offering a basis for a shared representational vocabulary. \citet{tennant_moral_2025} model different theories as reward functions in an iterated prisoner's dilemma setting, demonstrating that certain normative distinctions can be expressed computationally. However, these efforts are isolated and do not yet support standardized evaluation. A more systematic approach would involve constructing datasets that map normative theories to principles, rules, and reasoning patterns, with expert input and sufficient coverage to enable benchmarking.

The absence of shared ground truth also affects reliability. When benchmarks rely on different representations of the same theory, performance differences are difficult to interpret. A model may perform well under one operationalization of consequentialism and poorly under another, without a clear basis for comparison. In addition, temporal variation in human judgments \citep{keswani_moral_2025} introduces further instability when human annotations are used as reference points. Shared representations would not eliminate these issues but would provide a common basis for cross-benchmark comparison.

\subsection{Evaluating Reasoning Traces Without Normative Vocabulary}

A second limitation concerns the evaluation of reasoning traces. Even when benchmarks assess intermediate reasoning, they often lack the normative vocabulary needed to determine whether the norms invoked are appropriate, correctly applied, or properly weighted. MoReBench \citep{chiu_morebench_2025} illustrates this issue.

MoReBench evaluates reasoning using scenario-specific rubric items that combine theory-related and outcome-based criteria. Without a clear representation of how normative theories should be applied, it is difficult to distinguish between failures of norm understanding and failures of application.

The benchmark also uses a criterion-fulfillment scoring approach with length normalization to reduce verbosity bias. However, this creates incentives for minimal responses that satisfy rubric requirements without exposing reasoning, allowing less transparent models to achieve higher scores.

More generally, there is a disconnect between mentioning a morally relevant consideration and incorporating it into reasoning. Models may be penalized for implicit reasoning or rewarded for listing considerations without integrating them. While some work addresses aspects of this problem \citep{rao_ethical_2023}, current approaches remain limited. This is a broader challenge for evaluating reasoning processes, but it is particularly consequential in moral reasoning, where flawed metrics may misrepresent model capabilities.

\subsection{The Feature Problem}

A third limitation concerns the identification of morally relevant features. Before applying any norm, a system must determine which aspects of a situation are relevant for moral evaluation. Current approaches do not provide a general method for this step.

Existing work typically treats feature identification as task-specific. For example, \citet{kwon_neuro-symbolic_2024} generate features by prompting models to extract salient information across variations of a scenario. While effective in controlled settings, this approach does not generalize to novel situations. As noted in prior work, there is no unified computational account of how humans identify morally relevant features.

As a result, evaluations are constrained to the features anticipated by benchmark designers. This limits the ability to assess performance in settings where relevant considerations differ from those encoded in the dataset. The feature identification problem is closely related to the representation of normative theories, since many theories specify which aspects of a situation should be treated as morally relevant. Improved representations of normative principles would therefore support more general approaches to feature identification.

\section{Ways Forward}

We outline several directions for improving the evaluation of moral reasoning in LLMs.

\paragraph{Shared representations of normative theories.}
The field would benefit from common formal vocabularies that specify what different normative theories prescribe, including their principles, rules, and characteristic reasoning patterns. Prior work \citep{hammerton_fundamental_2025, tennant_moral_2025} provides initial steps, but existing efforts remain fragmented. Developing shared representations would enable the construction of standardized datasets linking theories to their endorsed norms, and would support comparability and aggregation of results across studies.

\paragraph{Expert-informed ground-truth data.}
Datasets for normative evaluation should incorporate input from domain experts across multiple ethical traditions such as consequentialism, deontology, virtue ethics, care ethics, and contractualism. Such datasets should be sufficiently large and structured to capture different levels of competence, including recognizing relevant norms, applying them in straightforward cases, and resolving conflicts between competing principles.

\paragraph{Separation of values-level and norms-level evaluation.}
Evaluations should explicitly distinguish between assessing value alignment (the moral value problem) and assessing norm application (the moral norm problem). This distinction should be reflected in both task design and reporting, rather than treated as a single construct.

\paragraph{Separation of deliberation and decision.}
The quality of a model's reasoning process and the correctness of its final judgment should be evaluated independently. A correct answer without appropriate reasoning does not demonstrate normative competence, and conversely, sound application of principles may yield non-standard conclusions. Conflating these dimensions obscures model capabilities.

\paragraph{Evaluation under assisted and unassisted settings.}
Benchmarks should assess both baseline performance (e.g., zero-shot responses) and performance under structured conditions, such as guided prompting, tool-usage, or multi-step deliberation. This enables evaluation of both observed and potential normative competence, allowing a clearer distinction between a model's inability to apply norms and a failure to elicit them.

\paragraph{Improved evaluation of reasoning traces.}
Current methods for assessing reasoning traces are limited. Future work should incorporate techniques from the chain-of-thought faithfulness literature to evaluate whether invoked norms contribute to final decisions~\citep{barez2025chain,swaroop2025frit}. In addition, evaluation methods using criterion-based scoring should distinguish between superficial mention of norms and their integration into reasoning.

\section{Conclusion}

The evaluation of moral competence in AI systems has made substantial progress on the moral value problem, i.e., whether models reflect human moral priorities. However, this focus has left the moral norm problem underexplored. Existing approaches, grounded in descriptive ethics, capture what models appear to value but do not assess whether they can apply normative principles to specific cases.

Addressing this limitation requires new evaluation infrastructure. In particular, the field needs shared representations of normative theories, expert-informed datasets that specify how norms apply across contexts, and evaluation protocols that distinguish between values-level alignment and norms-level reasoning. While these challenges are non-trivial, they are necessary for a complete assessment of moral competence. Evaluating what models care about is insufficient; it is equally important to evaluate whether they can determine what those commitments entail in practice.

\newpage

\bibliography{bibliography}

\end{document}